\documentclass[conference]{IEEEtran}
\IEEEoverridecommandlockouts
\usepackage{cite}
\usepackage{amsmath,amssymb,amsfonts}
\usepackage{algorithmic}
\usepackage{graphicx}

\usepackage{textcomp}
\usepackage{xcolor}
\usepackage{booktabs}
\usepackage{multirow}
\def\BibTeX{{\rm B\kern-.05em{\sc i\kern-.025em b}\kern-.08em
    T\kern-.1667em\lower.7ex\hbox{E}\kern-.125emX}}
\begin{document}

\title{Context Consistency Learning via Sentence Removal for Semi-Supervised Video Paragraph Grounding}



\author{
    \IEEEauthorblockN{
        Yaokun Zhong$^{1}$, 
        Siyu Jiang$^{4}$, 
        Jian Zhu$^{5}$,
        Jian-Fang Hu$^{1,2,3*}$
    }
    
    \IEEEauthorblockA{
        $^{1}$ School of Computer Science and Engineering, Sun Yat-sen University, China \\
        $^{2}$ Guangdong Province Key Laboratory of Information Security Technology, China \\
        $^{3}$ Key Laboratory of Machine Intelligence and Advanced Computing, Ministry of Education, China \\
        $^{4}$ Guangdong University of Foreign Studies   
        $^{5}$ Guangdong University of Technology \\
     Email: zhongyk23@mail2.sysu.edu.cn, jiangsiyu@gdufs.edu.cn, zj@gdut.edu.cn, hujf5@mail.sysu.edu.cn
    }

    \thanks{$^{*}$Corresponding author.}
}

\maketitle

\begin{abstract}
Semi-Supervised Video Paragraph Grounding (SSVPG) aims to localize multiple sentences in a paragraph from an untrimmed video with limited temporal annotations. Existing methods focus on teacher-student consistency learning and video-level contrastive loss, but they overlook the importance of perturbing query contexts to generate strong supervisory signals. In this work, we propose a novel Context Consistency Learning (CCL) framework that unifies the paradigms of consistency regularization and pseudo-labeling to enhance semi-supervised learning. Specifically, we first conduct teacher-student learning where the student model takes as inputs strongly-augmented samples with sentences removed and is enforced to learn from the adequately strong supervisory signals from the teacher model. Afterward, we conduct model retraining based on the generated pseudo labels, where the mutual agreement between the original and augmented views' predictions is utilized as the label confidence. Extensive experiments show that CCL outperforms existing methods by a large margin.
\end{abstract}


\section{Introduction}
\label{sec:intro}
In the area of multimodal machine learning, Video Grounding (VG)~\cite{tall, didemo, tan2021augmented, tan2022matching, tan2022context, lin2023collaborative, liang2025referdino} which aims to localize a given natural language query in an untrimmed video has become a popular and important problem. 
Although most of the past efforts are focused on the single-sentence setting of video grounding, Bao et al.~\cite{depnet} have recently extended the traditional single-sentence video grounding to a multi-sentence version called Video Paragraph Grounding (VPG), seeking better ways of utilizing the potential contextual information between multiple sentences to achieve more precise and unambiguous language-based temporal localization. Specifically, the task of VPG requires the model to simultaneously localize all the sentences in a given paragraph from an untrimmed video.

A series of related studies~\cite{depnet, prvg, svptr, hscnet} have been conducted on the problem of VPG, which has significantly promoted the development of fully-supervised methods. Despite the rapid progress of fully-supervised VPG models, the prohibitively expensive annotation cost of labeling precise temporal labels for sentence descriptions in untrimmed videos has limited the applications of fully-supervised models in more realistic scenarios. Recently, Jiang et al.~\cite{svptr} have proposed and explored the Semi-Supervised Video Paragraph Grounding (SSVPG) problem in order to relieve the heavy burden of acquiring manually annotated temporal labels. They introduced the first semi-supervised learning framework to address the problem of VPG, where they proposed a mean teacher framework that incorporates prediction consistency learning based on weak-strong augmentations and self-supervised learning with a video-level contrastive objective~\cite{xu2025efficient}. 

However, such a previously designed semi-supervised learning framework does not fully consider the task specificity of VPG, which makes it suffer from two severe defects that limit the learning performance of SSVPG models. First of all, this method adopts the common augmentation operations including feature channel shifting~\cite{tsm} and word masking~\cite{bert} to perturb the input samples for weak-strong teacher-student learning. Nevertheless, one of the essential aspects of addressing VPG lies in exploiting the contextual information, i.e., the temporal order and semantic relations across multiple sentences, to achieve dense events grounding in the video. The previously adopted augmentations are unable to destroy the query contexts across sentences, which means both the teacher and student can access almost the same amount of contextual information to make predictions with a similar degree of accuracy. This would significantly hinder the teacher from providing relatively strong enough supervision to the student. Besides, the previous method simply employs contrastive learning to distinguish video-paragraph pairs, while such video-level supervision suffers from a huge gap to suffice the moment-level localization.

In this work, we propose a novel Context Consistency Learning (CCL) framework to address the above drawbacks of current SSVPG approaches. Specifically, our method is designed as a two-stage framework to simultaneously benefit from consistency regularization and pseudo-labeling. To generate adequately strong supervision from teacher to student, we first develop a context-consistent mean teacher learning process, in which the query contexts of inputs are strongly perturbed by removing sentences from the raw paragraph. A contrastive-based consistency loss is then utilized to distill rich moment-level temporal supervision from teacher to the student. Furthermore, we infer the temporal pseudo labels of unlabeled samples and utilize the mutual agreement between model predictions across varying augmented query contexts as label confidence to boost performance by model retraining.

\begin{figure*}[t]
    \centering
    \includegraphics[width=1.0\linewidth]{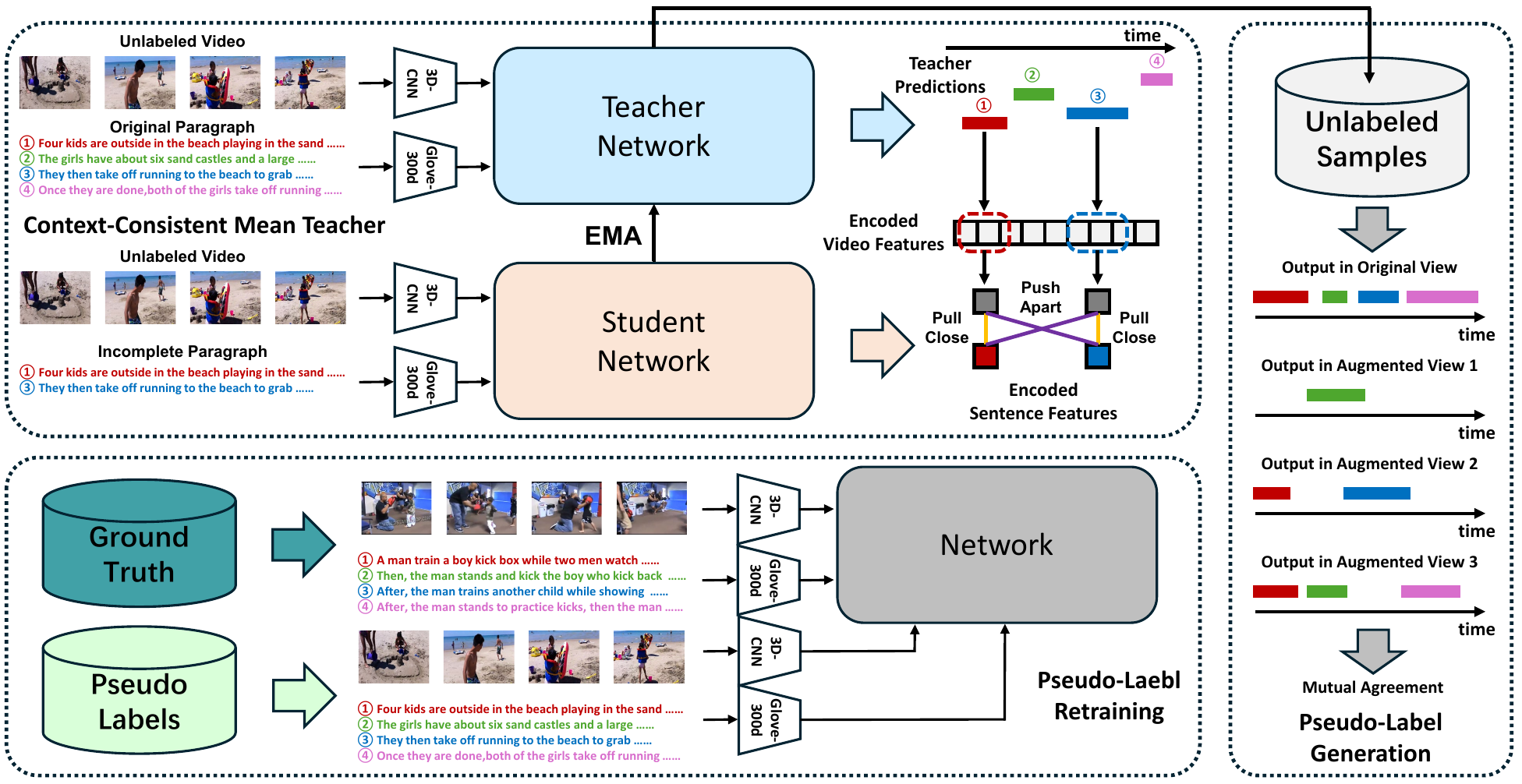}
    \vspace{-2em}
    \caption{An overview of our proposed context consistency learning method. It is a unified framework that conducts consistency regularization and pseudo-labeling in two stages for semi-supervised video paragraph grounding. In the first stage, the student takes as inputs strongly-augmented paragraphs generated by removing sentences and is enforced to learn a contrastive-based consistency loss from the teacher. In the second stage, we exploit the mutual agreement of model predictions under varying query contexts as label confidence to conduct pseudo-labeling retraining for better performance.}
    \label{fig:arch}
    \vspace{-1em}
\end{figure*}

\section{Related Work}

\noindent\textbf{Video Paragraph Grounding.} Bao et al.~\cite{depnet} have extended video grounding to a multi-query setting called video paragraph grounding, where the boundaries of multiple temporally ordered sentences in the paragraph are jointly localized from the video. 
A series of studies~\cite{depnet, svptr, hscnet, prvg, gtlr, tan2024synopground, SiamGTR} have also been conducted for this line of research. Bao et al.~\cite{depnet} proposed a Dense Events Propagation Network (DepNet) to adaptively aggregate the temporal and semantic information of multiple events into a compact set of representations and then selectively propagate useful information to each event. Shi et al.~\cite{prvg} proposed PRVG as the first end-to-end parallel decoding method for video paragraph grounding by re-purposing transformers. Tan et al.~\cite{hscnet} proposed a Hierarchical Semantic Correspondence Network (HSCNet) to learn the multi-level cross-modal semantic alignment and utilize dense supervision by grounding diverse levels of queries. Particularly, Jiang et al.~\cite{svptr} proposed a Semi-supervised Video-Paragraph TRansformer (SVPTR) and explored the semi-supervised learning of video paragraph grounding for the first time.

\noindent\textbf{Semi-Supervised Learning.} Semi-supervised learning aims to jointly exploit limited labeled data and abundant unlabeled data for better model performance. Basically, there exist two main paradigms in semi-supervised learning, including consistency regularization~\cite{mean-teacher, conmatch} that enforces consistent model outputs between different views of the same input and pseudo-labeling~\cite{pseudo-label, meta-pseudo-label, zhao2024weakly} that trains a model on limited labeled data and generates predictions on unlabeled data for further training. Jiang et al.~\cite{svptr} designed the first semi-supervised video paragraph grounding framework, which combines a typical mean teacher semi-supervised learning framework with self-supervised contrastive learning at the video-paragraph level. This method employs a simple language augmentation operation by masking out a proportion of the text words to generate supervision from the teacher to the model. However, this method cannot provide strong teacher-student supervision for VPG due to its preservation of contextual information between sentences. In contrast, we propose a novel semi-supervised framework to generate strong moment-level supervision from teacher to student using random sentence removal as an augmentation to perturb the query contexts.

\section{Methodology}
This section aims to illustrate our proposed framework to address video paragraph grounding under a semi-supervised setting. In the following, we will first give a problem definition of semi-supervised video paragraph grounding. Then, we elaborate on the design details of our method, including the model architecture, the two stages of context-consistent mean teacher training, and consistency-guided pseudo-label retraining.
\subsection{Problem Definition}
In this work, our goal is to boost the performance of semi-supervised video paragraph grounding by introducing a novel and more effective learning paradigm based on context consistency. Specifically, suppose an untrimmed video $\mathcal{V}$ and a paragraph description consisting of $N$ temporally ordered sentences, i.e., $\mathcal{P}=\{\mathcal{S}_i\}_{i=1}^{N}$ are given as the inputs, video paragraph grounding requires the model to simultaneously localize the temporal intervals $\mathcal{T}=\{(\tau_i^{\text{st}}, \tau_i^{\text{ed}})\}_{i=1}^{N}$ of dense events in the video described by the given sentences, where $\tau_i^{\text{st}}$ and $\tau_i^{\text{ed}}$ indicate the starting and ending time for the $i$-th sentence $\mathcal{S}_i$ in the paragraph, respectively. Different from the fully supervised setting that needs to acquire all the temporal annotations of the entire dataset, our focus is on the semi-supervised setting that only requires a small proportion of temporally labeled video-paragraph pairs, while utilizing the supervision provided by the massive amount of video-paragraph pairs without temporal labels.

\subsection{Context Consistency Learning}
\subsubsection{Overview} As shown in Figure~\ref{fig:arch}, our context consistency learning method seamlessly unifies the two primary paradigms in semi-supervised learning, i.e., consistency regularization and pseudo labeling for addressing semi-supervised video paragraph grounding. The entire training process of our proposed framework is divided into two stages. In the first stage, we develop a mean teacher framework using random sentence removal as the augmentation strategy for weak-strong consistency learning. A contrastive-based consistency loss is utilized to distill the more reliable moment-level temporal information from the teacher to the student. In the second stage, we generate pseudo labels on unlabeled data using the final teacher model of the first stage, during which the mutual agreement of model predictions from different contextual views of the same sample is regarded as an indicator of the label confidence. Combining the limited ground-truth labels and abundant pseudo labels with confidence, we retrain a network from scratch under a fully-supervised manner as our final model to achieve better generalization results. In the following, we will illustrate the details of our method.

\subsubsection{Context-Consistent Mean Teacher}

Traditional semi-supervised learning algorithms based on consistency regularization~\cite{mean-teacher, conmatch} directly employ a consistency constraint on model predictions between the teacher and student to generate supervisory signals. The existing semi-supervised video paragraph grounding method~\cite{svptr} adapts a classical mean teacher consistency learning framework by enforcing consistent temporal predictions. Differently, we develop a context-consistent mean teacher framework to produce strong teacher-to-student supervision and enhance fine-grained cross-modal representation learning.

\noindent\textbf{Feature Extraction.} To begin with, we  extract visual features $\mathbf{V}=\{\mathbf{v}_i\}_{i=1}^{T}\in \mathbb{R}^{T\times D_v}$ by pre-trained 3D-CNN~\cite{c3d}, and sentence-level query features $\mathbf{S} = \left\{\mathbf{s}_i\right\}_{i=1}^{N}\in \mathbb{R}^{N\times D_q}$ by Glove~\cite{glove} and Bidirectional Gated Recurrent Unit~\cite{gru}. Two linear projection layers are employed to project the visual and query features into a common embedding space to obtain $\mathbf{F}_v\in \mathbb{R}^{T\times D}$ and $\mathbf{F}^{'}_q\in \mathbb{R}^{N\times D}$ as the input of teacher model.

\noindent\textbf{Sentence Removal Perturbation.} Different from the previous works~\cite{stlg, svptr} that mainly conducts input perturbations at the channel dimension while maintaining the sentence contexts that queries can access almost unaffected, we completely remove some sentences from the query contexts as the input perturbation operation. Specifically, $M$ sentences are randomly removed from the original set of sentences in the paragraph, which forms a contextually strongly-augmented input as:
\begin{equation}
    \mathbf{F}_q \leftarrow \left\{\mathbf{F}^{'}_q\left(i\right) \mid \mathcal{S}_i \notin \Omega, \Omega \subset \mathcal{P}\right\},
\end{equation}
where $\mathbf{F}_q\in \mathbb{R}^{(N - M)\times D}$ denotes the query features augmented with sentence removal and $\Omega$ denotes the $M$ sentences randomly selected from the original paragraph. Compared to those naive feature perturbations like shifting channel activations or adding gaussian noise, random sentence removal can increase sample diversity by enhancing contextual variations and enforce the model to learn better cross-modal event reasoning abilities that are crucial for video paragraph grounding.

\noindent\textbf{Mean Teacher Self-Training.} As presented in Figure~\ref{fig:arch}, we maintain a teacher model and a student model that have exactly the same encoder-decoder network structure. During training, the student model is updated by minimizing the loss functions with gradient descent, while the teacher model is not updated with backpropagation but uses an Exponential Moving Average (EMA) of weights from the student model. 

Following the previous works~\cite{detr, prvg}, we adopt a transformer-based network architecture to conduct language-conditioned parallel regression for video paragraph grounding. It mainly consists of a transformer encoder to encode the complex temporal structures of input videos and a transformer decoder to decode the localization results of each sentence by absorbing relevant information from the video modality. Specifically, denote the parameters of the teacher and student model as $\theta^{'}$ and $\theta$, respectively, then the teacher model is updated as the following:
\begin{equation}
    \theta^{'}_t = \gamma\theta^{'}_{t-1} + (1 - \gamma)\theta_{t-1},
\end{equation}
where the subscript $t$ represents the number of training iterations and $\gamma$ is a momentum coefficient to control the smoothness of the parameter update process. In the training phase, the teacher model is used to provide stable learning targets~\cite{mean-teacher} to facilitate learning of the student model. To achieve this goal, we first input the video and query features to the teacher model and obtain the predicted temporal intervals $\hat{\mathcal{T}}^{'} = \{(\hat{\tau}_i^{\text{st}}, \hat{\tau}_i^{\text{ed}})\}_{i=1}^{N}$ for all given sentences.
Then the pseudo temporal labels generated by the teacher model to guide the online learning of the student model are represented as:
\begin{equation}
    \mathcal{T}^{'} = \left\{\hat{\mathcal{T}}^{'}_{\sigma\left(i\right)} \mid i \in \left[1, N-M \right] \right\},
\end{equation}
where $\sigma\left(i\right)$ indicates an index mapping between teacher and student predictions. Since the student model is fed with a strongly-augmented query input with sentences removed, the teacher model that receives the original inputs can produce more precise boundary predictions than the student model. 

However, typically a large amount of teacher predictions still cannot be very close to the ground truth, which tends to inject considerable noise into the student model learning when prediction consistency is employed as the objective. To conduct effective self-training based on the mean teacher framework, we utilize a contrastive learning objective to enable more effective weak-strong semi-supervised learning. Concretely, we first obtain the encoded video features $\mathbf{V}_{enc}$ in the student model from $\mathbf{F}_v$, then a set of moment-level visual features are constructed as:
\begin{equation}
    \mathbf{F}_m\left(i\right) = \frac{\sum_{j}{\beta_{i, j} \mathbf{V}_{enc}\left(j\right)}}{\sum_j \beta_{i, j}},
\end{equation}
where $\mathbf{F}_m\in \mathbb{R}^{(N-M)\times D}$ denotes the moment-level features corresponding to each sentence in the augmented query input. $\beta_{i, j}$ takes 1 if the $j$-th clip falls within the temporal interval of the $i$-th sentence predicted by the teacher, otherwise 0. Thereafter, we employ a contrastive loss as follows:
\begin{align}
    \mathcal{L}_{con} &= \frac{1}{N-M}\sum_{i=1}^{N-M}
    \left(
    \frac{\text{exp}\left(\text{cos}\left(\mathbf{F}_m(i), \mathbf{F}_q(i) \right) / \tau \right)}{\sum_{j} \text{exp}\left(\text{cos}\left(\mathbf{F}_m(i), \mathbf{F}_q(j) \right) / \tau \right)}
    \right) \nonumber\\
    &+ \frac{1}{N-M}\sum_{i=1}^{N-M}
    \left(
    \frac{\text{exp}\left(\text{cos}\left(\mathbf{F}_m(i), 
    \mathbf{F}_q(i) \right) / \tau \right)}{\sum_{j} \text{exp}\left(\text{cos}\left(\mathbf{F}_m(j), \mathbf{F}_q(i) \right) / \tau \right)}
    \right),
    \label{eq:contrastive_loss}
\end{align}
where $\tau$ is the temperature hyper-parameter to control the sharpness of the contrastive loss. By contrasting the moment-level features and sentence-level features under the guidance of the teacher model's predictions, the student model can more softly learn the consistency supervision and reduce the adverse impact brought by inaccurate teacher labels.

\subsubsection{Consistency-Guided Pseudo-Labeling}
In our semi-supervised method, the model is first trained by the context-level weak-strong consistency regularization, during which it learns to better localize dense events in the video indicated by various query contexts under the guidance of relatively more precise temporal predictions from the teacher model. Despite the effectiveness of consistency regularization, existing methods like FixMatch~\cite{fixmatch} have shown that it is possible to unify the two primary semi-supervised paradigms, i.e., consistency regularization and pseudo-labeling to achieve a more satisfactory learning performance.

In this work, we explore a way to seamlessly integrate pseudo-labeling into our holistic framework based on context consistency between model predictions to achieve better results. We hypothesize that the model shows a high confidence level when it makes highly consistent predictions on the same sample with varying sentence contexts, i.e., the mutual agreement between predictions of multiple augmented views generated by sentence removal tends to indicate the confidence level of the pseudo labels. Formally, the model's context consistency for a given sample can be formulated as:
\begin{equation}
    C = \frac{1}{N-1}\sum_{k=1}^{N-1} \frac{1}{k}\sum_{j=1}^{k} \text{IoU}\left(\hat{\mathcal{T}}^{a}_{k, j}, \hat{\mathcal{T}}^{o}_{\sigma\left(j\right)} \right),
    \label{eq: context consistency}
\end{equation}
where $\hat{\mathcal{T}}^{a}_{k}\in \mathbb{R}^{k\times 2}$ denotes the model prediction of the augmented input in which $N-k$ sentences are randomly removed and $\hat{\mathcal{T}}^{o}\in \mathbb{R}^{N\times 2}$ denotes the model prediction of the original input. $\sigma\left(j\right)$ indicates an index mapping that finds the prediction in $\hat{\mathcal{T}}^{o}$ corresponding to a prediction in $\hat{\mathcal{T}}^{a}_k$. 

Based on Equation (\ref{eq: context consistency}), we develop a simple yet effective mechanism to utilize context consistency as guidance to enhance the self-training. Concretely, the pseudo labels are first divided into three groups according to their context consistency. The pseudo labels with low consistency are excluded from the label set, and the high-consistency pseudo labels are learned with larger loss weights while the mid-consistency pseudo labels are learned with relatively smaller loss weights.

\begin{table*}[t]
    \centering
    \footnotesize
    \caption{Comparison with state-of-the-arts on ActivityNet-Captions, Charade-CD-OOD and TACoS datasets.}
    \vspace{-1em}
    \begin{tabular}{l|c|cccc|cccc|cccc}
    \toprule[1pt]
    \multirow{2}*{Method} & \multirow{2}*{Setting}& \multicolumn{4}{c}{ActivityNet-Captions} \vline & \multicolumn{4}{c}{Charades-CD-OOD} \vline & \multicolumn{4}{c}{TACos}\\
    \cmidrule(lr){3-6} \cmidrule(lr){7-10} \cmidrule(lr){11-14}
    & & R@0.3 & R@0.5 & R@0.7 & mIoU & R@0.3 & R@0.5 & R@0.7 & mIoU & R@0.3 & R@0.5 & R@0.7 & mIoU\\
    \midrule[1pt]
    3D-TPN~\cite{2dtan} & FS & 67.56 & 51.49 & 30.92 & - & - & - & - & - & 55.05 & 40.31 & 26.54 & -\\
    DepNet~\cite{depnet} & FS & 72.81 & 55.91 & 33.46 & - & 45.61 & 27.59 & 10.69 & 29.30 & 56.10 & 41.34 & 27.16 & - \\
    STLG~\cite{stlg} & FS & - & - & - & - & 48.30 & 30.39 & 9.79 & - & - & - & - & -\\
    PRVG~\cite{prvg} & FS & 78.27 & 61.15 & 37.83 & 55.62 & - & - & - & - & 61.64 & 45.40 & 26.37 & 29.18\\
    SVPTR~\cite{svptr} & FS & 78.07 & 61.70 & 38.36 & 55.91 & 55.14 & 32.44 & 15.53 & 36.01 & 67.91 & 47.89 & 28.22 & 31.42\\
    HSCNet~\cite{hscnet} & FS & 81.89 & 66.57 & 44.03 & 59.71 & - & - & - & - & 76.28 & 59.74 & 42.00 & 40.61\\
    \midrule[1pt]
    DepNet~\cite{depnet} & SS & 61.46 & 45.14 & 26.78 & 44.11 & 43.03 & 25.07 & 10.14 & 28.09 & 40.27 & 26.95 & 16.54 & 18.68\\
    STLG~\cite{stlg} & SS & - & - & - & - & 46.15 & 29.43 & 9.38 & - & - & - & - & -\\
    VPTR~\cite{svptr} & SS & 72.80 & 53.14 & 29.07 & 50.08 & 45.13 & 24.98 & 10.22 & 28.92 & 61.31 & 40.59 & 21.39 & 26.59 \\
    SVPTR~\cite{svptr} & SS & 73.39 & 56.72 & 32.78 & 51.98 & 50.31 & 28.50 & 12.27 & 32.13 & 63.06 & 40.19 & 20.05 & 26.10 \\
    CCL (Ours) & SS & \textbf{79.58} & \textbf{62.23} & \textbf{36.61} & \textbf{56.00} & \textbf{57.24} & \textbf{32.33} & \textbf{14.59} & \textbf{37.24} & \textbf{68.25} & \textbf{43.42} & \textbf{23.59} & \textbf{28.79} \\
    \midrule[1pt]
    \end{tabular}
    \label{tab: comp_activity_charades}
    \vspace{-1em}
\end{table*}

\subsection{Training Details}
The training process of our proposed context consistency learning framework can be divided into the mean teacher self-training stage and the pseudo-label retraining stage. 

In the first stage, we employ a location loss $\mathcal{L}_{loc}$ and an attention loss $\mathcal{L}_{att}$ on the labeled samples and a contrastive-based consistency loss $\mathcal{L}_{con}$ is applied to the unlabeled samples. The location loss $\mathcal{L}_{loc}$ consists of an L1 loss and a GIoU loss~\cite{giou}, and $\mathcal{L}_{att}$~\cite{mun2020local} is an attention-guided loss on each decoder layer.
The training loss for the mean teacher self-training stage is given as:
\begin{equation}
    \mathcal{L}_{ts} = \lambda_1\left(\mathcal{L}_{loc} + \mathcal{L}_{att} \right) + \lambda_2\mathcal{L}_{con},
\end{equation}
where $\lambda_1$ and $\lambda_2$ are loss weights to balance the contributions of the two terms. 

For the retraining stage, we employ both location loss and attention loss on the ground-truth labels, high-consistency pseudo labels and mid-consistency pseudo labels. Note that the low-consistency pseudo labels are not used for training. The total loss for pseudo-label retraining can be given as:
\begin{equation}
    \mathcal{L}_{rt} = \lambda_3\left(\mathcal{L}^{gt}_{loc} + \mathcal{L}^{gt}_{att} \right) + \lambda_4\left(\mathcal{L}^{hc}_{loc} + \mathcal{L}^{hc}_{att} \right) + \lambda_5\left(\mathcal{L}^{mc}_{loc} + \mathcal{L}^{mc}_{att} \right),
\end{equation}
where $\left(\mathcal{L}^{gt}_{loc}, \mathcal{L}^{gt}_{att}\right)$, $\left(\mathcal{L}^{hc}_{loc}, \mathcal{L}^{hc}_{att}\right)$, $\left(\mathcal{L}^{mc}_{loc}, \mathcal{L}^{mc}_{att}\right)$ denote the loss terms for the ground-truth labels, high-/mid-consistency pseudo labels, respectively. $\lambda_3$, $\lambda_4$ and $\lambda_5$ are loss weights to balance the three different kinds of supervisory signals.


\section{Experiments}
\label{sec:experiment}
\subsection{Datasets and Metrics}
\noindent\textbf{ActivityNet-Captions.} ActivityNet-Captions~\cite{activitynet_captions} is a large-scale open-domain video dataset containing 14,926 videos and 71,953 moment-sentence pairs. On average, each video has a total duration of 117.60 seconds and each annotated paragraph consists of 3.63 sentences. The entire dataset is divided into train/val\_1/val\_2 sets comprised of 10,009/4,917/4,885 video-paragraph pairs, respectively. Following the setting adopted by previous works~\cite{depnet, svptr, prvg, hscnet}, we use the val\_2 set for testing.

\noindent\textbf{Charades-CD-OOD.} Charades-STA~\cite{tall} is a video grounding dataset focusing on a wide range of indoor activities. For a fair comparison, we adopt the Charades-CD-OOD splits proposed in~\cite{closerlook} for evaluation. It is divided into train/val/test\_ood sets consisting of 4,564/333/1,440 video-paragraph pairs, respectively. The average video length is 30.78 seconds and the average paragraph length is 2.41 sentences.

\noindent\textbf{TACoS.} TACoS~\cite{tacos} is a dataset for video grounding which is sourced from the cooking videos of MPII corpus~\cite{MPII}. There are 127 long untrimmed videos and each video is manually annotated with multiple paragraphs at different descriptive granularities. The training, validation, and testing sets in TACoS consist of 1,107, 418, and 380 video-paragraph pairs, respectively. The average video length and number of sentences per paragraph are 4.79 minutes and 8.75, respectively.

\noindent\textbf{Evaluation Metrics.} We adopt the mean Intersection over Union (i.e., mIoU) and recall values calculated by a certain IoU threshold of $m$ (i.e., $\text{R@m}$) for evaluation metrics, as commonly used in the existing methods~\cite{depnet, svptr}. $m$ is set to be $\{0.3, 0.5, 0.7\}$ for ActivityNet-Captions and Charades-CD-OOD, and $\{0.1, 0.3, 0.5\}$ for TACoS.

\subsection{Implementation Details}
In all experiments, we adopt the same C3D network~\cite{c3d} and Glove model~\cite{glove} as feature extractors for a fair comparison~\cite{depnet, prvg}. We implement a three-layer transformer encoder and a three-layer transformer decoder with cosine similarity instead of dot product, following~\cite{hscnet}. The model hidden size $D$ is set to $256$ in all experiments. The number of sampled video clips $T$ is set to be $256$, $128$, and $512$ for ActivityNet-Captions, Charades-CD-OOD, and TACoS datasets, respectively. We train the model using Adam optimizer with a fixed learning rate of $0.0001$ and a batch size of $32$, $32$, and $16$ for ActivityNet-Captions, Charades-CD-OOD and TACoS, respectively. The temperature hyper-parameter for $\mathcal{L}_{con}$ is $0.01$ in all settings. The loss weights $\left\{\lambda_1, \lambda_3, \lambda_4, \lambda_5\right\}$ are set as $\left\{2, 2, 4, 2\right\}$ and $\lambda_2$ is set to $0.75$, $0.75$, $1.5$ on ActivityNet-Captions, Charades-CD-OOD and TACoS, respectively.

\subsection{Performance Comparison}
To verify the effectiveness of our proposed method for semi-supervised learning in video paragraph grounding, we conduct extensive experiments on all of the three publicly available datasets used in previous works~\cite{depnet, svptr, prvg, hscnet}, i.e., ActivityNet-Captions~\cite{activitynet_captions}, Charades-CD-OOD~\cite{tall} and TACoS~\cite{tacos} to report the quantitative results. Specifically, we compare our approach with other Semi-Supervised (SS) methods, including DepNet~\cite{depnet}, STLG~\cite{stlg}, VPTR~\cite{svptr} and SVPTR~\cite{svptr}. In addition, we also report quantitative results of the state-of-the-art Fully-Supervised (FS) methods. 

As shown in Table~\ref{tab: comp_activity_charades}, the quantitative results demonstrate that our proposed CCL method significantly outperforms all of the existing state-of-the-art methods under the same semi-supervised setting on the three datasets.
Concretely, our method achieves a gain of $4.02\%$, $5.11\%$ and $2.69\%$ in terms of mIoU on ActivityNet-Captions, Charades-CD-OOD and TACoS datasets, respectively. Notably, our semi-supervised method can achieve comparable results with some FS methods, further demonstrating our method's superiority.

\subsection{Ablation Study}
In this section, we aim to quantitatively study the contributions of different components in our proposed method by conducting a series of ablation experiments, and the results are shown in Table~\ref{tab: main_ablation}. First, we can see that using the exponential moving average of model weights can bring certain performance improvements, i.e., $0.2\%$ mIoU by comparing row 1 and row 3. Then, it can be observed that directly employing random sentence augmentation on labeled data for training would not work well comparing row 1 and row 2. For the two primary designs of our proposed context consistency learning, i.e., the context-consistent contrastive regularization and consistency-guided pseudo-labeling can both improve the model performance by a considerable margin. Concretely, for row 3 and row 5, we can see that adopting the contrastive-based consistency loss on the teacher-student model gives a $1.03\%$ gain in mIoU. For the strictest metric R@0.7, our consistency learning even brings a higher improvement up to $1.95\%$, which shows the efficacy of generating strong moment-level supervision for the student model. Last but not least, we find that further conducting model retraining based on the consistency-guided pseudo labels can boost the model performance by $0.49\%$ mIoU.
\begin{table}[t]
    \centering
    \caption{Ablation studies on the main components of our proposed CCL framework. "MT", "Aug.", "CR", "PL" are short for the mean teacher framework, sentence removal augmentation, consistency regularization and pseudo-labeling in our framework, respectively.}
    \vspace{-1em}
    \begin{tabular}{cccc|cccc}
    \toprule[1pt]
    MT & Aug. & CR & PL & R@0.3 & R@0.5 & R@0.7 & mIoU \\
    \toprule[1pt]
    \checkmark & & & & \small{77.76} & \small{58.91} & \small{34.64} & \small{54.28} \\
    & \checkmark & & & \small{77.70} & \small{58.90} & \small{33.15} & \small{53.96} \\
    \checkmark & \checkmark & & & \small{78.96} & \small{59.78} & \small{33.92} & \small{54.48} \\
    \checkmark & & \checkmark & & \small{78.45} & \small{58.64} & \small{32.68} & \small{53.97} \\
    \checkmark & \checkmark & \checkmark & & \small{79.60} & \small{61.89} & \small{35.87} & \small{55.51} \\
    \checkmark & \checkmark & \checkmark & \checkmark & \small{79.58} & \small{62.23} & \small{36.61} & \small{56.00} \\
    \toprule[1pt]
    \end{tabular}
    \label{tab: main_ablation}
    \vspace{-2em}
\end{table}

\subsection{Qualitative Results}
\begin{figure}[h]
    \centering
    \includegraphics[width=\linewidth]{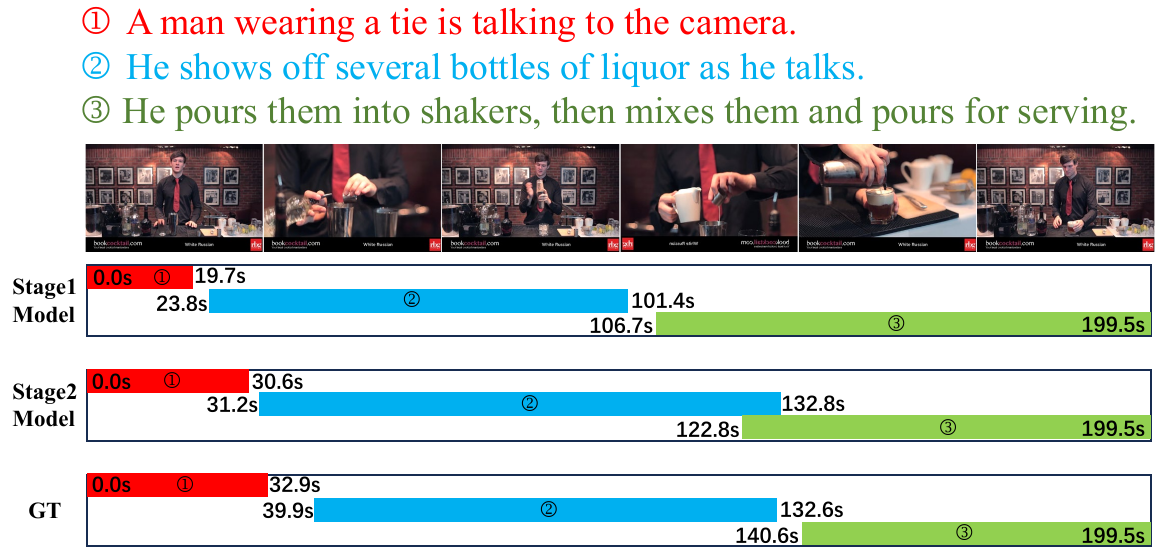}
    \vspace{-2em}
    \caption{Qualitative visualization on model predictions of our framework.}
    \label{fig:visual}
    \vspace{-1em}
\end{figure}

Here we give an intuitive visualization of the model predictions obtained in the two stages of our proposed framework. As shown in Figure~\ref{fig:visual}, the first row shows the model prediction
from the teacher model in our first stage’s consistency learning, which can roughly locate the positions of three events in the video. Furthermore, we visualize the model prediction from the second stage. It can be clearly observed that this model prediction is more accurate than the former one, showing the effectiveness of our proposed pseudo-label retraining process.
\vspace{-2em}

\section{Conclusion}
In this work, we propose a novel Context Consistency Learning (CCL) framework to address the Semi-Supervised Video Paragraph Grounding (SSVPG) problem. Our proposed CCL framework is the first semi-supervised learning framework that unifies the consistency regularization and pseudo-labeling paradigms for SSVPG. It includes two stages of mean teacher self-training and pseudo-label model retraining. Experiments show that our method significantly surpasses the previous state-of-the-art methods.

\vspace{1mm}
\begin{sloppypar}
{
{\noindent\textbf{Acknowledgements.} This work was partially supported by NSFC (U21A20471, 62476296), Guangdong Natural Science Funds Project (2023B1515040025, 2022B1515020009), Guangdong Provincial Science and Technology Program Project (2024A1111120017), and Guangdong Provincial Key Laboratory of Information Security Technology (2023B1212060026).}}
\end{sloppypar}

\bibliographystyle{IEEEbib}
\bibliography{ICME25}

\end{document}